\documentclass[10pt,twocolumn,letterpaper]{article}

\usepackage{amsmath,amsfonts,bm}

\def\eqref#1{equation~\ref{#1}}

\def\1{\bm{1}}

\DeclareMathAlphabet{\mathsfit}{\encodingdefault}{\sfdefault}{m}{sl}
\SetMathAlphabet{\mathsfit}{bold}{\encodingdefault}{\sfdefault}{bx}{n}

\usepackage{cvpr}              

\usepackage{graphicx}
\usepackage{amsmath}
\usepackage{amssymb}
\usepackage{booktabs}

\usepackage{url}
\usepackage{enumitem}
\usepackage{makecell}
\usepackage{diagbox}
\usepackage{bm}
\usepackage{dsfont}
\usepackage{multirow}
\usepackage{varwidth}
\usepackage{pifont}
\usepackage{makecell}
\usepackage{wrapfig}
\usepackage{graphicx} 
\usepackage{comment}
\usepackage{color}
\usepackage{xcolor}
\usepackage{colortbl,booktabs}

\usepackage{algorithmic}
\usepackage[ruled, linesnumbered]{algorithm2e}

\usepackage{xspace}
\usepackage{booktabs}
\usepackage{rotating}
\usepackage{color}

\newcommand{\cmark}{\ding{51}\xspace}%
\newcommand{\xmark}{\ding{55}\xspace}%
\newcommand{\xmarkg}{\textcolor{lightgray}{\ding{55}}\xspace}%

\definecolor{Tianlong_color}{rgb}{0.858, 0.188, 0.478}

\usepackage[pagebackref,breaklinks,colorlinks]{hyperref}

\usepackage[capitalize]{cleveref}
\crefname{section}{Sec.}{Secs.}
\Crefname{section}{Section}{Sections}
\Crefname{table}{Table}{Tables}
\crefname{table}{Tab.}{Tabs.}

\def\cvprPaperID{4325} 
\def\confName{CVPR}
\def\confYear{2022}

\begin{document}

\title{The Principle of Diversity: Training Stronger Vision Transformers Calls for Reducing All Levels of Redundancy}

\author{%
  Tianlong Chen\textsuperscript{1}, Zhenyu Zhang\textsuperscript{1}, Yu Cheng\textsuperscript{2}, Ahmed Awadallah\textsuperscript{2}, Zhangyang Wang\textsuperscript{1}\\
  \textsuperscript{1}University of Texas at Austin, \textsuperscript{2}Microsoft Research \\
  \small{\texttt{\{tianlong.chen, zhenyu.zhang, atlaswang\}@utexas.edu}, \texttt{\{yu.cheng, hassanam\}@microsoft.com}} \\
}

\maketitle

\begin{abstract}
Vision transformers (ViTs) have gained increasing popularity as they are commonly believed to own higher modeling capacity and representation flexibility, than traditional convolutional networks. However, it is questionable whether such potential has been fully unleashed in practice, as the learned ViTs often suffer from over-smoothening, yielding likely redundant models. Recent works made preliminary attempts to identify and alleviate such redundancy, e.g., via regularizing embedding similarity or re-injecting convolution-like structures. However, a ``head-to-toe assessment" regarding the extent of redundancy in ViTs, and how much we could gain by thoroughly mitigating such, has been absent for this field. This paper, for the first time, systematically studies the ubiquitous existence of redundancy \underline{at all three levels}: patch embedding, attention map, and weight space. In view of them, we advocate a \underline{principle of diversity} for training ViTs, by presenting corresponding regularizers that encourage the representation diversity and coverage at each of those levels, that enabling capturing more discriminative information. Extensive experiments on ImageNet with a number of ViT backbones validate the effectiveness of our proposals, largely eliminating the observed ViT redundancy and significantly boosting the model generalization. For example, our diversified DeiT obtains $0.70\%\sim1.76\%$ accuracy boosts on ImageNet with highly reduced similarity. Our codes are fully available in {\small \url{https://github.com/VITA-Group/Diverse-ViT}}.
\vspace{-1em}
\end{abstract}


\vspace{-3mm}
\section{Introduction}
\vspace{-1mm}
Transformer~\cite{vaswani2017attention}, as the \textit{de facto} neural architecture  in natural language processing (NLP)~\cite{devlin2018bert,brown2020language}, recently revolutionizes modern computer vision applications such as image classification~\cite{dosovitskiy2020image,touvron2020training,han2020survey}, object detection~\cite{zheng2020end,carion2020end,dai2020up,zhu2021deformable}, and image generation~\cite{parmar2018image,pmlr-v119-chen20s,jiang2021transgan}. Rather than relying on convolution-like inductive bias, vision transformers~\cite{dosovitskiy2020image} (ViTs) leverage the self-attention~\cite{vaswani2017attention} to aggregate image patches across all spatial positions and model their global-range relationships, which are believed to improve model expressiveness and representation flexibility.

\begin{figure}[t]
    \centering
    \includegraphics[width=0.9\linewidth]{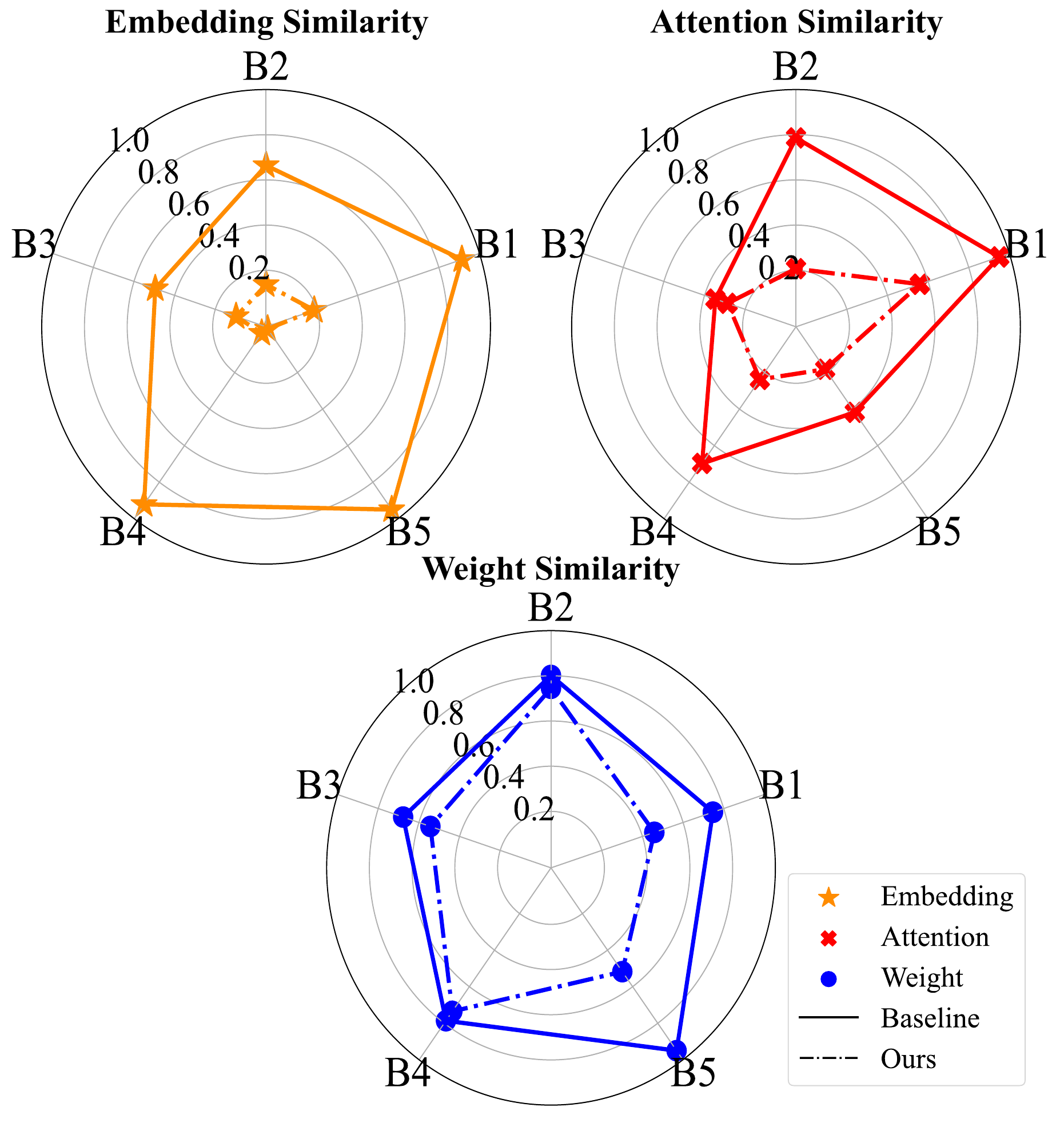}
    \vspace{-0.5em}
    \caption{Relative similarity comparisons in embedding, attention, and weight spaces of DeiT-Small on ImageNet. The larger number indicates severer correlation/redundancy. B1$\sim$B5 donate the blocks in the DeiT-Small model. \textit{Cosine}, (normalized) \textit{MSE}, $1$ - (normalized) \textit{reconstruction loss} are adopted to measure embedding, attention, and weight similarity. The former two are computed with $10,000$ images from the ImageNet training set without data augmentation, following the standard in~\cite{gong2021vision}. 
    }
    \vspace{-6mm}
    \label{fig:teaser}
\end{figure}

Despite their promising potentials, the ViT training still suffers from considerable instability, especially when going deeper~\cite{touvron2021going,gong2021vision}. One of the major reasons~\cite{gong2021vision} is that the global information aggregation among all patches encourages their representations to become overly similar, causing substantially degraded discrimination ability. This phenomenon, known as over-smoothening, suggests a high degree of ``redundancy" or ineffective usage of the ViT expressiveness and flexibility, and has been studied by a few prior arts~\cite{touvron2021going,zhou2021deepvit,zhou2021refiner,gong2021vision}. Several initial attempts strive to fill the gap from different aspects. For example \cite{gong2021vision} proposes contrastive-based regularization to diversity patch embeddings, and \cite{zhou2021refiner} directly refines the self-attention maps via convolution-like aggregation to augment local patterns. 


This paper aims to comprehensively study and mitigate the ViT redundancy issue. We first systematically demonstrate the ubiquitous existence of redundancy \textbf{at all three levels}: \textit{patch embedding, attention map, and weight space}, for current state-of-the-art (SOTA) ViTs. That is even the case for those equipped with strong data augmentations (i.e., DeiT~\cite{touvron2021training}) or sophisticated attention mechanisms (i.e., Swin~\cite{liu2021swin}), e.g, as shown in Figure~\ref{fig:teaser}. 
In view of such collapse, we advocate a \textbf{principle of diversity} for training ViTs, by proposing corresponding regularizers that encourage the representation diversity and coverage at each of those levels, that unleashes the true discriminative power and representation flexibility of ViTs. We find each level's regularizers to provide generalization gains, and applying them altogether consistently yields superior performance. 
Our contributions lie in the following aspects: \vspace{-0.3em}
\begin{itemize}
    \item We provide the first comprehensive investigation of redundancy in ViTs by demonstrating its ubiquitous existence in all three levels of patch embeddings, attentions, and weights, across SOTA ViT models.
    \vspace{-0.2em}
    \item For each of the three levels, we present diversity regularizers for training ViTs, which demonstrate complementary effects in eliminating redundancy, encouraging diversity, and enhancing generalization. \vspace{-0.2em} 
    \item We conduct extensive experiments with vanilla ViT, DeiT, and Swin transformer backbones on the ImageNet datasets, showing consistent and significant performance boost gains by addressing the tri-level redundancy issues with our proposed regularizers. Specifically, our proposals improve DeiT and Swin, by $0.70\%\sim 1.76\%$ and $0.15\% \sim 0.32\%$ accuracy. \vspace{-0.3em}
\end{itemize}

\section{Related Works}

\paragraph{Vision transformer.} Transformer~\cite{vaswani2017attention} emerges from NLP applications with prevailing successes, motivating its adaptation to the computer vision scenarios~\cite{dosovitskiy2020image}. ViTs encodes an image into a sequence of patches and feeds them to transformer encoders. Such self-attention based models get rid of common inductive biases, e.g., locality in convolutional neural networks (CNNs), and the global interactions among embedding grant ViTs a stronger learning capacity. ViT's empirical successes in various computer vision tasks include image classification~\cite{dosovitskiy2020image,guo2021cmt,chen2021chasing}, object detection~\cite{zheng2020end,carion2020end,dai2020up,zhu2021deformable}, segmentation~\cite{wang2020max,wang2020end,zhao2020point},  enhancement~\cite{chen2020pre,yang2020learning}, image generation~\cite{parmar2018image,pmlr-v119-chen20s,jiang2021transgan}, video and vision-language understanding~\cite{zeng2020learning,zhou2018end,lu2019vilbert,tan2019lxmert,chen2020uniter,su2019vl,li2019visualbert,li2020unicoder,li2020oscar,zhou2020unified}.   

However, the global information aggregation of ViTs also results in over-smoothed and redundant representations~\cite{gong2021vision,zhou2021refiner}. That makes the effective learning capacity of ViTs ``collapsed", which prohibits ViTs from practically achieving higher capacity and sophisticated representations.
Existing works have taken two angles: ($i$) (re-)injecting locality via convolution-like structures and fusing global and local contexts, \underline{for self-attention}~\cite{zhou2021refiner,han2021transformer,liu2021swin,wu2021cvt,xiao2018unified,jiang2021all,valanarasu2021medical,arnab2021vivit,touvron2021going,xie2021so,yuan2021tokens}; ($ii$) adopting patch-wise contrastive or mixing loss to boost  diversity, \underline{for patch embeddings}. Aiming to connect and expand those isolated efforts, our work is the first to target the full-scale redundancy in ViTs at embedding, attention, and weight levels. Note that our tri-level diversification framework is compatible with existing approaches by plugging them in the corresponding level. Detailed investigations are presented in Section~\ref{sec:ablation}.



\vspace{-1em}
\paragraph{Diversity regularization.} Diversity constraints are designed to learn discriminative patterns for improved feature coverage and generalization~\cite{gong2021vision,Chen_2019_ICCV,liu2021learning}. Representative regularizers include cosine similarity-based ~\cite{gong2021vision}, the margin or distance-based \cite{schroff2015facenet,liu2016large,NIPS2016_6b180037,chen2020simple,khosla2020supervised,gong2021vision}, the hyperspherical uniformity-based~\cite{liu2018learning,lin2020regularizing,liu2021learning}, and the orthogonality-based~\cite{bansal2018can,Chen_2019_ICCV,huang2018orthogonal,arjovsky2016unitary,vorontsov2017orthogonality,rodriguez2016regularizing,lee2019orthogonality,8922788,zhang-etal-2021-orthogonality}. Most of them are applid to CNNs, with \cite{gong2021vision} making the recent attempt in ViTs.


\section{Methodology} \label{sec:methods}
\begin{figure*}[t]
    \centering
    \includegraphics[width=1.0\linewidth]{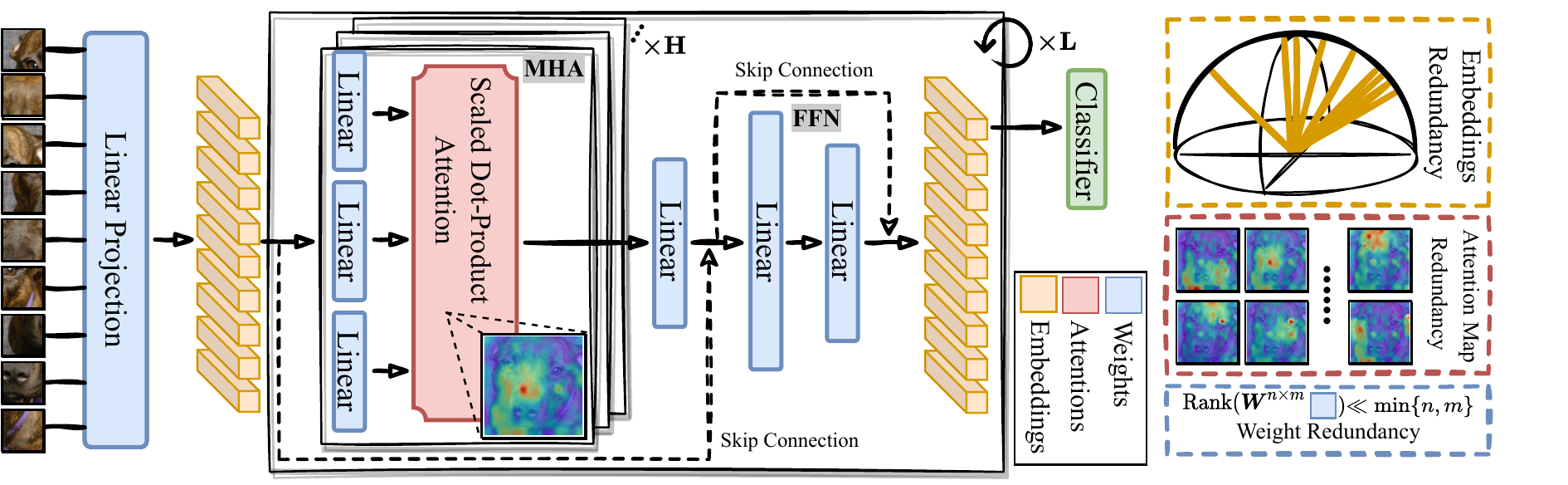}
    \vspace{-8mm}
    \caption{(\textit{Left}) A overall pipeline of vision transformers~\cite{dosovitskiy2020image,touvron2021training}. Each image is divided into patches and transformed into embeddings via a linear projection layer. Then, embeddings are fed to the transformer encoder consists of MHA and FFN modules. Other operations like softmax and normalization are omitted here. (\textit{Right}) An illustration of the redundancy of embedding, attention, and weight.}
    \vspace{-3mm}
    \label{fig:DViT}
\end{figure*}

\subsection{Examining the Tri-Level Redundancy in ViTs}
\paragraph{Preliminaries.} Revisit that transformer architectures~\cite{vaswani2017attention,dosovitskiy2020image} usually contain the multi-head self-attention modules (MHA) and feed-forward networks (FFN). In MHA, keys, queries, and values are linearly transformed for computing the attention heads, and then all the heads are aggregated by another linear transformation. FFNs are also built on two linear transformations with activations, as shown in Fig.~\ref{fig:DViT}. 

Here, we use $\boldsymbol{W}^{\mathrm{MHA}}$ and $\boldsymbol{W}^{\mathrm{FFN}}$ to denote the weights in MHA and FFN modules, respectively. $\boldsymbol{A}$ represents the attention map (or the affinity matrix). It is calculated by $\boldsymbol{A}=\mathrm{softmax}(\alpha\boldsymbol{Q}\boldsymbol{K}^{\top})$, where $\boldsymbol{Q}$ is the query matrix, $\boldsymbol{K}$ is the key matrix, and $\alpha$ is a scale (typically $\frac{1}{\sqrt{d}}$ and $d$ is the dimension of the keys and the queries). Let $\boldsymbol{e}^{l}=[\boldsymbol{e}^{l}_{\mathrm{class}},\boldsymbol{e}^{l}_1,\cdots,\boldsymbol{e}^{l}_n]$ be the feature embedding of layer $l$ ($1\le l\le \mathrm{L}$), where $n$ is the total number of image patches. Without loss of generality, we take image recognition as an example. Then, the vision transformer is optimized by minimizing a classification loss $\mathcal{L}(\mathcal{C}(\boldsymbol{e}_{\mathrm{class}}^{\mathrm{L}}),y)$, where $\mathcal{C}$ is the classification head and $y$ is the label of input samples. 

\vspace{-1em}
\paragraph{Redundancy in patch embeddings.} We investigate the redundancy of the feature embedding by calculating the token-wise cosine similarity. It is depicted as follows:
\begin{equation}
\mathcal{R}_{\mathrm{cosine}}^{s}(\boldsymbol{h}):=\frac{1}{n(n-1)}\sum_{i\not = j}\frac{|h_i^{^\top}h_j|}{\|h_i\|_2\|h_j\|_2},
\label{eq:cosine_same}
\end{equation}
\begin{equation}
\mathcal{R}_{\mathrm{cosine}}^{d}(\boldsymbol{h}^{l_1},\boldsymbol{h}^{l_2}):=\frac{1}{n}\sum_{i}\frac{|h_i^{l_1^\top}h_i^{l_2}|}{\|h_i^{l_1}\|_2\|h_i^{l_2}\|_2},
\label{eq:cosine_different}
\end{equation}
where $\boldsymbol{h}$ is the feature embedding $\boldsymbol{e}=[\boldsymbol{e}_{\mathrm{class}},\boldsymbol{e}_1,\cdots,\boldsymbol{e}_n]$ (superscript $l$ is omitted for simplicity), and $n$ is the total number of tokens.

Notably, $\mathcal{R}_{\mathrm{cosine}}^{s}(\boldsymbol{h})$ and $\mathcal{R}_{\mathrm{cosine}}^{d}(\boldsymbol{h}^{l_1},\boldsymbol{h}^{l_2})$ denote the cosine similarity of the feature embedding within the \underline{s}ame layer and across two \underline{d}ifferent layers $l_1$, $l_2$, respectively. The larger cosine similarity suggests more redundancy. Intuitively, the within-layer redundancy hinders ViT from capturing different tokens' features; and the cross-layer redundancy hurts the learning capacity of ViTs since highly correlated representations actually collapse the effective depth of ViT to fewer or even single transformer layer. 

\paragraph{Redundancy in attentions.} We consider $\mathcal{R}_{\mathrm{cosine}}(\boldsymbol{A})$ to measure the cosine similarity of attention maps within the same layer. Similarly, $\mathcal{R}_{\mathrm{MSE}}(\boldsymbol{A}):=\frac{1}{n(n-1)}\sum_{i\not = j}\|A_i-A_j\|_2^2$ can also be used for the redundancy quantification. In contrast to these two metrics which show the similarity across attention heads, we further use the standard deviation statistics to indicate the element-wise variance within an attention head. 

\paragraph{Redundancy in model weights.} If the parameter space is highly redundant, then the weight matrix will fall approximately into a low-rank parameters subspace. Thus, we use the reconstruction error to depict the weight redundancy:
\begin{equation}
    \mathcal{R}_{\mathrm{PCA}}(\boldsymbol{W}):=\|\boldsymbol{W} - \Tilde{\boldsymbol{W}}\|_2^2
\end{equation}
where $\Tilde{\boldsymbol{W}}$ is the reconstructed weight matrix by the principal component analysis (PCA) with the top-$k$ principal components. Given a fixed reconstruction error, the larger $k$ implies better diversity. In other words, given $k$, the larger reconstruction error means less weight redundancy. \cite{Chen_2019_ICCV,liu2021learning} also dissect the weight redundancy from the view of rank.

\subsection{Eliminating the Tri-Level Redundancy in ViTs}
To mitigate the observed redundancy, we introduce three groups of regularization to encourage the diversity of $i)$ learned feature embeddings; $ii)$ attention maps; $iii)$ model weights in the training of vision transformers.

\vspace{-2mm}
\paragraph{Patch embedding diversity.} To diversify patch feature embeddings, we use the cosine angle regularization $\mathcal{R}_{\mathrm{cosine}}^{s}(\boldsymbol{e})$ and $\mathcal{R}_{\mathrm{cosine}}^{d}(\boldsymbol{e}^{l_1},\boldsymbol{e}^{l_2})$ to constrain within-layer and cross-layer embedding, respectively. Similar methods are leveraged to obtain diversified representations in vision~\cite{gong2021vision}, language~\cite{gao2019representation}, and graph~\cite{chen2020measuring}  scenarios. Meanwhile, we adopt the contrastive regularization $\mathcal{R}_{\mathrm{contrastive}}^{d}(\boldsymbol{e}^{l_1},\boldsymbol{e}^{l_2})$ to boost cross-layer embedding diversity, which is presented as follows:
\begin{equation}
\begin{aligned}
    &\mathcal{R}_{\mathrm{contrastive}}(\boldsymbol{e}^{l_1}, \boldsymbol{e}^{l_2}) := \\
    &-\frac{1}{n}\sum_{i=1}^n \mathrm{log} \frac{\mathrm{exp}(e_i^{l_1^\top} e_i^{l_2})}{\mathrm{exp}(e_i^{l_1^\top} e_i^{l_2}) + \mathrm{exp}(e_i^{l_1^\top}(\frac{1}{n-1}\sum_{j\not=i}e_j^{l_2}))},
\label{eq:contrastive}
\end{aligned}
\end{equation}
where $l_1$ and $l_2$ are two different layer indexes. Note that the contrasitve regularizer is not applicable for the within-layer embedding diversification since the lack of positive pairs. 

\noindent $\rhd$ \underline{\textit{Rationale.}} As pointed out by~\cite{merikoski,gong2021vision}, the cosine angle regularization can function like minimizing the upper bound of the largest eigenvalue of patch embedding $\boldsymbol{e}$, hence bringing improvements of expressiveness~\cite{gong2021vision} and diversity to learned representations. For contrastive regularization, it pulls embeddings corresponding to the same patch together and simultaneously pushes apart embeddings belonging to different patches, reducing the feature correlation between different layers. As a result, it enables to learn separable patch embedding and maintain tolerance to semantically similar patches~\cite{wang2020understanding,wang2021understanding}, improving the representation qualities and the ViT performance.


\paragraph{Attention diversity.} In the same way, the cosine regularization $\mathcal{R}^s_{\mathrm{cosine}}(\boldsymbol{A})$ can be applied to remove the redundancy of attention, where $\boldsymbol{A}=[\boldsymbol{A}_1,\boldsymbol{A}_2,\cdots,\boldsymbol{A}_{\mathrm{H}}]$ and $\mathrm{H}$ is the number of attention heads within one layer. Inspired by the orthogonality regularization's empirical effectiveness in vision~\cite{Chen_2019_ICCV,lezama2018ole,ranasinghe2021orthogonal} and language tasks~\cite{zhang-etal-2021-orthogonality}, we investigate it under the context of ViTs. We adopt the canonical soft orthogonal regularization (SO)~\cite{bansal2018can} as follows:
\begin{equation}
    \mathcal{R}_{\mathrm{SO}}(\boldsymbol{A}):=\|\boldsymbol{A}^\top \boldsymbol{A} - \boldsymbol{I}\|_\mathrm{F}^2,
\end{equation}
where $\|\cdot\|_\mathrm{F}$ is the Frobenius norm and $\boldsymbol{I}$ is the identity matrix sharing the same size as $\boldsymbol{A}^\top \boldsymbol{A}$.

We also try an alternative Conditional number orthogonal regularization (CNO) \cite{Chen_2019_ICCV} as follows:
\begin{equation}
    \mathcal{R}_{\mathrm{CNO}}(\boldsymbol{A}) = \| \lambda_1(\boldsymbol{A}^\top\boldsymbol{A})-\lambda_2(\boldsymbol{A}^\top\boldsymbol{A})\|^2.
\end{equation}
It enforces the orthogonality via directly regularizing the conditional number $\kappa=\frac{\lambda_1}{\lambda_2}$ to 1, where $\lambda_1$ and $\lambda_2$ are the largest and smallest eigenvalues of the target matrix $\boldsymbol{A}^\top\boldsymbol{A}$. To make it computationally more tractable and stable, we alternatively constrain the difference between $\lambda_1$ and $\lambda_2$.

\noindent $\rhd$ \underline{\textit{Rationale.}} These regularizations (i.e., SO and CNO) encourage diverse attention maps by constraining them to be orthogonal with each other, which actually upper-bounds the Lipschitz constant of learned function mappings~\cite{zhang-etal-2021-orthogonality}, leading to robust and informative representations. As illustrated in~\cite{zhang-etal-2021-orthogonality}, introducing an orthogonal diversity regularizer to the attention map also stabilizes the transformer training and boosts its generalization on NLP tasks. 

\paragraph{Weight diversity.} Similarly, the orthogonality regularization, e.g., $\mathcal{R}_{\mathrm{CNO}}(\boldsymbol{W})$, can be easily plugged in and promote the diversity in ViT's weight space. Compared to orthogonality, hyperspherical uniformity is another more general diversity regularization demonstrated in~\cite{liu2021learning}. Although it has been explored in CNNs, its study in ViTs has been absent so far. We study the minimum hyperspherical separation (MHS) regularizer, which maximizes the separation distance (or the smallest pairwise distance) as follows:
\begin{equation}
    \mathrm{max}_{\{\hat{\boldsymbol{w}}_1,\cdots,\hat{\boldsymbol{w}}_m\}\in\mathbb{S}^{t-1}}\{\mathcal{R}_{\mathrm{MHS}}(\hat{\boldsymbol{W}}):=\mathrm{min}_{i\not= j}\rho(\hat{\boldsymbol{w}}_i,\hat{\boldsymbol{w}}_j)\},
\label{eq:mhs}
\end{equation}
where $\boldsymbol{W}=[\boldsymbol{w}_1,\boldsymbol{w}_2\cdots,\boldsymbol{w}_m]$, $\hat{\boldsymbol{w}_i}:=\frac{\boldsymbol{w}_i}{\|\boldsymbol{w}_i\|}$ is the $i$th weight vector projected onto a unit hypersphere $\mathbb{S}^{t-1}:=\{\hat{\boldsymbol{w}}\in\mathbb{R}^t|\|\hat{\boldsymbol{w}}\|=1\}$, $\rho(\cdot,\cdot)$ is the geodesic distance on the unit hypersphere. As indicated in Equation~\ref{eq:mhs}, it is formulated as a max-min optimization and we solve it with alternative gradient ascent/descent. 

Furthermore, we examine another maximum gram determinant (MGD) regularizer $\mathcal{R}_{\mathrm{MGD}}(\hat{\boldsymbol{W}})$ as follows:
\begin{equation}
    \mathrm{max}_{\{\hat{\boldsymbol{w}}_1,\cdots,\hat{\boldsymbol{w}}_m\}\in\mathbb{S}^{t-1}}\mathrm{logdet}\big(\boldsymbol{G}:=(\mathcal{K}(\hat{\boldsymbol{w}}_i,\hat{\boldsymbol{w}}_j))^m_{i,j=1}\big),
\label{eq:mgd}
\end{equation}
where $\mathrm{det}(\boldsymbol{G})$ is the determinant of the kernel gram matrix $\boldsymbol{G}\in\mathbb{R}^{m\times m}$ and $\mathcal{K}(\boldsymbol{u},\boldsymbol{v}):=\mathrm{exp}(-\sum_{i=1}^t\epsilon^2(u_i-v_i)^2)$ denotes the kernel function with a scale $\epsilon>0$. By maximizing the $\mathrm{det}(\boldsymbol{G})$ of weights $\hat{\boldsymbol{W}}$, MGD forces weight vectors to uniformly dispersed over the hypersphere.    

\noindent $\rhd$ \underline{\textit{Rationale.}} As demonstrated in~\cite{liu2018learning,lin2020regularizing,liu2021learning}, the hyperspherical uniformity regularizations (i.e., MHS and MGD) characterizes the diversity of vectors on a unit hypersphere, which encodes a strong inductive bias with relational information. We believe it to benefit ViT training from two perspectives~\cite{liu2021learning}: ($i$) eliminating weight redundancy and improving the representative capacity; ($ii$) learning better optimization and generalization by reducing the spurious local minima, evidenced in~\cite{liu2021learning,xie2017diverse,lin2020regularizing,liu2018learning}. 

\begin{figure*}[t]
    \centering
    \includegraphics[width=1.0\linewidth]{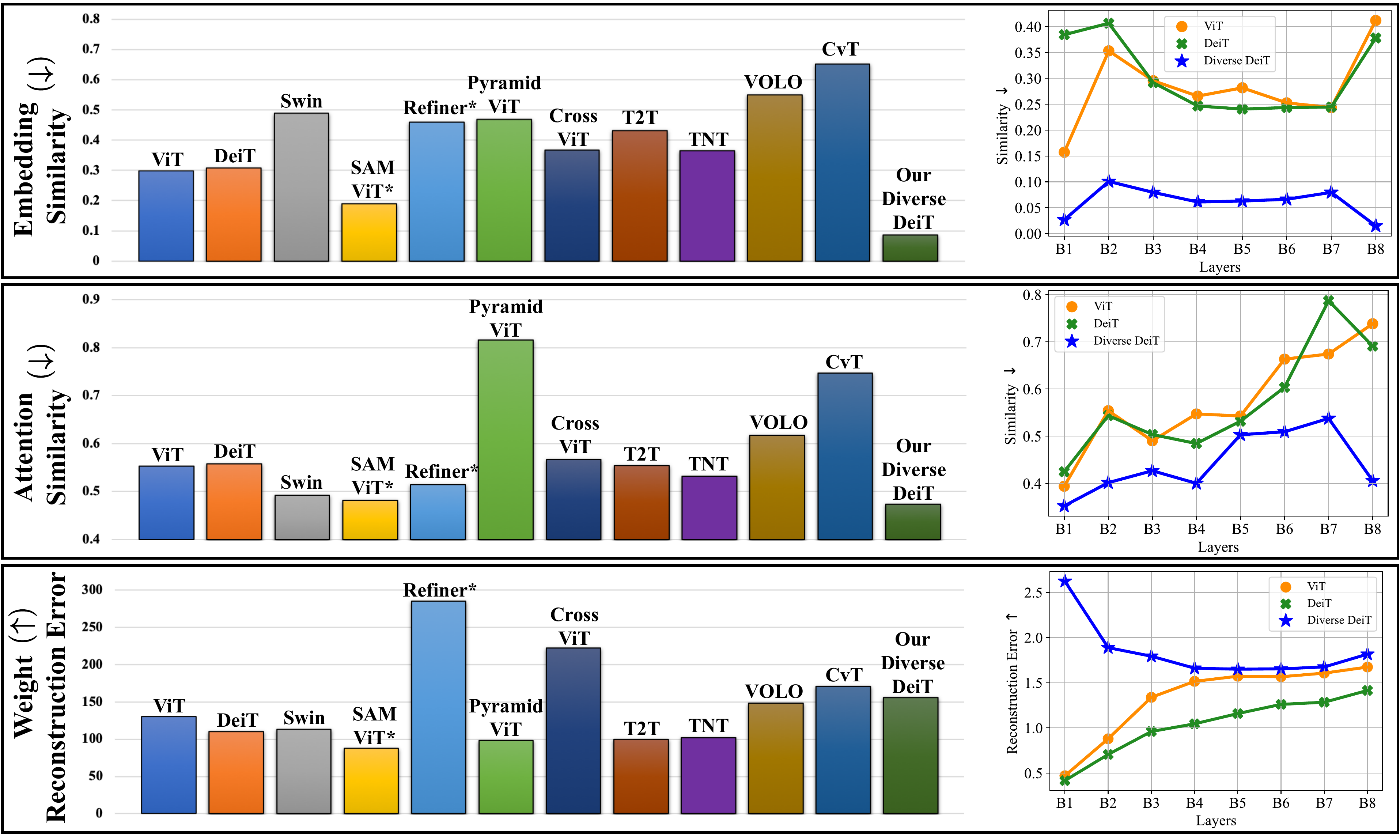}
    \vspace{-8mm}
    \caption{(\textit{Left}) Redundancy comparisons in embedding, attention, and weight spaces of ViT~\cite{dosovitskiy2020image}, DeiT~\cite{touvron2020training}, Swin~\cite{dosovitskiy2020image}, SAM-ViT~\cite{chen2021vision}, Refiner~\cite{zhou2021refiner}, Pyramid-ViT~\cite{wang2021pyramid}, Cross-ViT~\cite{chen2021crossvit}, T2T~\cite{yuan2021tokens}, TNT~\cite{han2021transformer}, VOLO~\cite{yuan2021volo}, CvT~\cite{wu2021cvt}, and our diverse DeiT on ImageNet. We use publicly available pre-trained models for benchmarking their all levels of redundancy. For a fair comparison, most of selected pre-trained transformers share similar parameter counts, i.e., $19\mathrm{M}\sim27\mathrm{M}$, while even the smallest released SAM-ViT and Refiner have $83$M and $78$M. $\uparrow$/$\downarrow$ denote that the larger/smaller number indicates better diversity. \textit{Cosine}, \textit{cosine}, (normalized) \textit{reconstruction error} are adopted to measure embedding, attention, and weight similarity. The former two are computed with $10,000$ sub-sampled images from the ImageNet training set without data augmentation, following the standard in~\cite{gong2021vision}. (\textit{Right}) The layer-wise similarity/reconstruction error of ViT, DeiT, and Swin, which B1$\sim$B8 is the corresponding transformer blocks (or layers).}
    \vspace{-3mm}
    \label{fig:current_vit}
\end{figure*}

\section{Experiment}

\paragraph{Implementation details.} We conduct extensive experiments on the ImageNet-1k~\cite{deng2009imagenet} dataset with ViT~\cite{dosovitskiy2020image}, DeiT~\cite{touvron2020training} and Swin transformers~\cite{liu2021swin}. All the hyperparameters of our introduced diversity regularizations are carefully tuned by a grid search, and the best configurations are provided in Section~\ref{sec:more_implementation}. Tesla V100-SXM2-32GB GPUs are used as our computing resources. Specifically, each experiment is ran with $8$ V100s for $1\sim 4$ days.

For vanilla ViT models, we consider two architectures with $12$ layers, i.e., ViT-Small and ViT-Base, which contains $6$ and $12$ heads for the multi-head self-attention block in each layer, respectively. We train each model for $300$ epochs with a batch size of $4096$. An AdamW optimizer is adopted with $0.3$ weight decay, and the learning rate starts from $4\times10^{-3}$ with $4$ epochs for warm-up and decays by a cosine annealing schedule. We maintain all training settings the same as the original ones in~\cite{dosovitskiy2020image}.

For DeiT architectures, we choose DeiT-Small, DeiT-Small24, and DeiT-Base. Specifically, both DeiT-Small and DeiT-Base contain $12$ layers, while DeiT-Small24 has $24$ layers. In each layer, DeiT-Small and DeiT-Small24 have $6$ heads for the self-attention module, and DeiT-Base has $12$ heads. Following~\cite{touvron2020training}, we train the models for $300$ epochs with the batch size of $1024$. We use an AdamW optimizer of $0.05$ weight decay. The initial learning rate is $1\times10^{-3}$ with $5$ epochs of warm-up, which reduces by a cosine annealing schedule. More details about data augmentation and other training tricks can be found in~\cite{touvron2020training}, 

For Swin transformer, we start from the official Swin-Small and Swin-Base pre-trained models and then fine-tune them for another $30$ epochs, in which we use a constant $1\times10^{-5}$ learning rate, $1\times10^{-8}$ weight decay and a batch size of $1024$~\cite{gong2021vision}. We keep other training details the same as~\cite{liu2021swin} and compare the fine-tuning performance with/without diversity regularizations.

\subsection{Redundancy in Current ViTs} 
In this section, we conduct a thorough investigation to reveal the broadly existing redundancy in ViT's patch embedding, attention map and weight spaces. Specifically, $11$ current SOTA ViTs and our diversified DeiT are examined on ImageNet, which can be grouped into four categories according to their proposed approaches on ViT~\cite{dosovitskiy2020image}: ($i$) improving training techniques such as data augmentation in DeiT~\cite{touvron2020training} and flatness-aware regularizer in SAM-ViT~\cite{chen2021vision}; ($ii$) introducing convolution layers or crafting convolution-like operations like CvT~\cite{wu2021cvt} and Refiner~\cite{zhou2021refiner} (termed as CNN + ViT); ($iii$) designing hierarchical structures to capture multi-scale information such as Swin~\cite{dosovitskiy2020image}, Pyramid-ViT~\cite{wang2021pyramid}, and Cross-ViT~\cite{chen2021crossvit}; ($iv$) exploring finer-level features or encoding local contexts into featuring like T2T~\cite{yuan2021tokens}, TNT~\cite{han2021transformer}, and VOLO~\cite{yuan2021volo}. From Figure~\ref{fig:current_vit}, we observe that:

\begin{figure*}[t]
    \centering
    \includegraphics[width=1.0\linewidth]{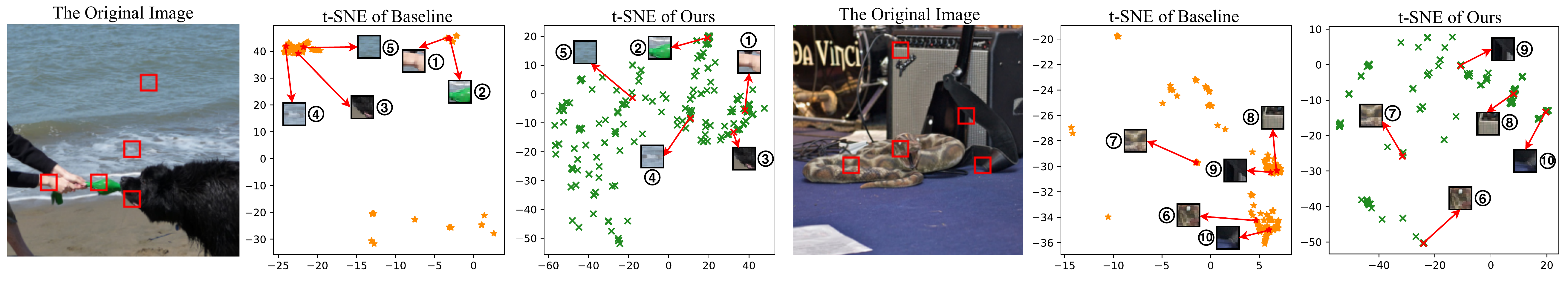}
    \vspace{-8mm}
    \caption{t-SNE visualization for patch embedding (i.e., $196$ tokens) of images randomly sampled from ImageNet. The patch embedding in \{$2$nd, $5$th\}/\{$3$rd, $6$th\} columns are generated from the Deit baseline and our diversified variant, respectively. Patches \ding{172}$\sim$\ding{181} are corresponding to the red box in the original images.}
    \vspace{-3mm}
    \label{fig:tSNE_feature}
\end{figure*}


\begin{figure*}[t]
    \centering
    \includegraphics[width=1\linewidth]{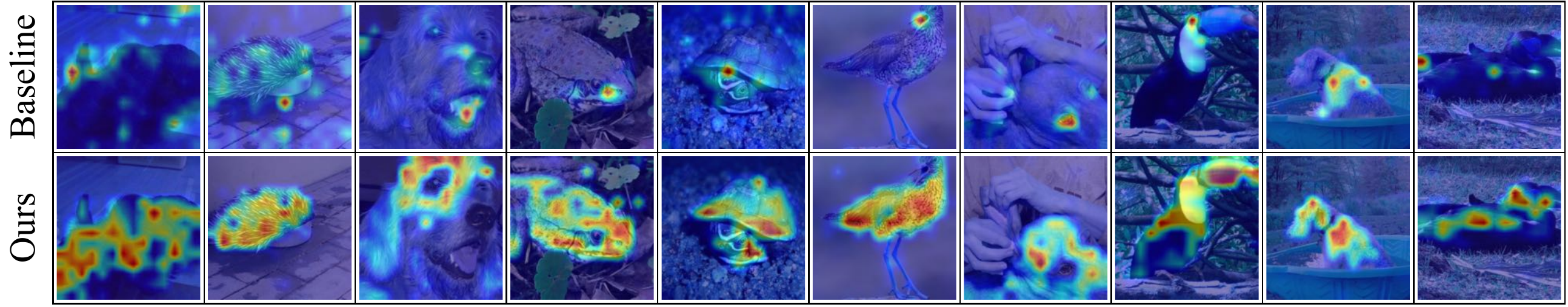}
    \vspace{-7mm}
    \caption{Attention visualizations of DeiT on ImageNet. Heatmaps in the $1$st and $2$nd rows are from baseline and our diversified variant. Tools of interpreting ViTs in~\cite{chefer2021transformer} are adopted to produce visualization.}
    \vspace{-4mm}
    \label{fig:gradcam_attention}
\end{figure*}

\vspace{-1mm}
\begin{itemize}
    \item [\ding{182}] At the patch embedding level, most of current SOTA ViTs exclude SAM-ViT have amplified embedding redundancy compared to vanilla ViT. It seems to suggest that the flatness-aware regularizer can enable ViT to generate more diverse embeddings. Furthermore, our diverse DeiT presents a consistent and substantial reduction on both layer-wise and overall embedding similarity, outperforming all other ViT variants. \vspace{-1mm}
    \item [\ding{183}] At the attention level, our diverse DeiT again achieves superior diverse attention maps with the least redundancy. In addition, self-attention with shifted windows in Swin~\cite{dosovitskiy2020image} and convolution-like aggregation in Refiner~\cite{zhou2021refiner} effectively reduce the attention map correlation. Also, flatness-aware SAM-ViT brings more diversity across attention heads. \vspace{-1mm}
    \item [\ding{184}] At the weight level, with similar parameter counts, Cross-ViT~\cite{chen2021crossvit} with a dual-branch structure for extracting multi-scale features has the largest weight reconstruction error, implying more weight diversity. Besides, our diverse DeiT still consistently outperforms its vanilla counterpart as shown in Figure~\ref{fig:current_vit} (\textit{right}), which validates our proposal's effectiveness in eliminating the weight space redundancy.
\end{itemize}

\begin{table}[t]
\centering
\caption{Top-1 accuracies (\%) of ViTs with or without all-level diversification on ImageNet. $\uparrow$ indicates the performance gains from our diversity regularizers compared with its vanilla version. All ViT and DeiT transformers are trained from scratch for $300$ epochs with the default setup in~\cite{dosovitskiy2020image} and~\cite{touvron2020training}. Swin Transformers are fine-tuned from released checkpoints from~\cite{liu2021swin} for $30$ epochs.}
\label{tab:sota}
\vspace{-3mm}
\scalebox{0.85}
{\small
\begin{tabular}{ll|cc}
\toprule
\multicolumn{2}{l|}{\scalebox{1}{Settings \& Methods}} & \multicolumn{1}{c}{Size} & \multicolumn{1}{c}{Accuracy} \\ \midrule
\multirow{2}{*}{(CNN)} & ResNet-152~\cite{he2016deep} & 230M & 78.10 \\ 
 & DenseNet-201~\cite{huang2017densely} & 77M & 77.60 \\
\midrule
(CNN + ViT) & CVT-21~\cite{wu2021cvt} & 32M & 82.50 \\
(ViT + DiversePatch$^\clubsuit$) & DeiT-Small~\cite{gong2021vision} & 22M & 80.43\\
(ViT + DLA$^\spadesuit$) & Refiner~\cite{zhou2021refiner} & 86M & 81.20\\
\midrule
\multirow{4}{*}{(Vanilla ViT $12$ layers)} & ViT-Small~\cite{dosovitskiy2020image} & 22M & 76.54 \\
& ViT-Small + Ours & 22M & 78.60 \scalebox{0.8}{($\uparrow2.04$)} \\
& ViT-Base~\cite{dosovitskiy2020image} & 86M & 77.90 \\
& ViT-Base + Ours & 86M & 79.96 \scalebox{0.8}{($\uparrow2.06$)} \\
\midrule
\multirow{4}{*}{(DeiT $12$ layers)}  & DeiT-Small~\cite{touvron2020training} & 22M & 79.78\\
& DeiT-Small + Ours & 22M & 80.61 \scalebox{0.8}{($\uparrow0.83$)}\\
& DeiT-Base~\cite{touvron2020training} & 86M & 80.98 \\
& DeiT-Base + Ours & 86M & 81.68 \scalebox{0.8}{($\uparrow0.70$)} \\
\midrule
\multirow{4}{*}{(Swin $12$ layers)} & Swin-Small~\cite{liu2021swin} & 50M & 83.18 \\
& Swin-Small + Ours & 50M & 83.33 \scalebox{0.8}{($\uparrow0.15$)}\\
& Swin-Base~\cite{liu2021swin} & 88M & 83.40 \\
& Swin-Base + Ours & 88M & 83.72 \scalebox{0.8}{($\uparrow0.32$)}\\
\midrule
\multirow{2}{*}{(DeiT $24$ layers)} & DeiT-Small24~\cite{touvron2020training} & 43M & 80.03 \\
 & DeiT-Small24 + Ours & 43M & 81.79 \scalebox{0.8}{($\uparrow1.76$)} \\
\bottomrule
\end{tabular}}
\footnotesize{\leftline{$\clubsuit$ DiversePatch is the diversification on the token embedding level in~\cite{gong2021vision}.}} \\ 
\footnotesize{\leftline{$\spadesuit$ DLA is the diversification on the attention level in~\cite{zhou2021refiner}.}}
\vspace{-5mm}
\end{table}

\subsection{Enhanced ViTs by Introducing Diversity}

\paragraph{Superior generalization of our proposal.} In this section, we demonstrate that improving diversity in ViT training achieves superior generalization. In specific, our experiments consider six representative transformer backbones, i.e., vanilla ViT-Small/Base, DeiT-Small/Base and Swin-Small/Base. As shown in Table~\ref{tab:sota}, several consistent observations can be drawn: \ding{182} Compared to ViT, DeiT\footnote{We disable the repeated augmentation in DeiT-Base's~\cite{touvron2021training} training schemes due to the well-known loss NAN issue of the original implementation (\url{https://github.com/facebookresearch/deit/issues/29}), which will result in slight performance degradation.} and Swin baselines, our diversified variants obtain $\sim2\%$, $0.70\%\sim 1.76\%$, and $0.15\%\sim 0.32\%$ accuracy boosts respectively, which evidence the effectiveness of our tri-level diversity regularization. \ding{183} Vanilla ViTs tend to benefit more from our diversity-ware training, by $3\sim7$ times improvements on DeiT and Swin. A possible explanation is that the data augmentation in DeiT and self-attention with shifted windows in Swin have already injected a certain level of diversity, as suggested by the reduced attention and weight correlation in Figure~\ref{fig:current_vit}. \ding{184} Compared with existing methods on the same ViT backbone, our diverse ViTs gain $0.18\%$ and $0.48\%$ accuracy improvements over DiversePatch~\cite{gong2021vision} and DLA~\cite{zhou2021refiner} respectively, which indicate the necessity of reducing all levels of redundancy. \ding{185} Deep ViTs receive more benefits from diversity regularized training. Particularly, DeiT-Small24 achieves a $1.76\%$ accuracy increase, while its $12$-layer variants have $0.83\%$ accuracy gains. It's within expectation since deeper transformers usually suffer harsher over-smoothing and representation redundancy~\cite{dong2021attention,zhou2020bert}, leaving more potential advancement for diversification. Similar conclusions can be observed in~\cite{gong2021vision}. 

\begin{figure*}[!ht]
    \centering
    \includegraphics[width=0.93\linewidth]{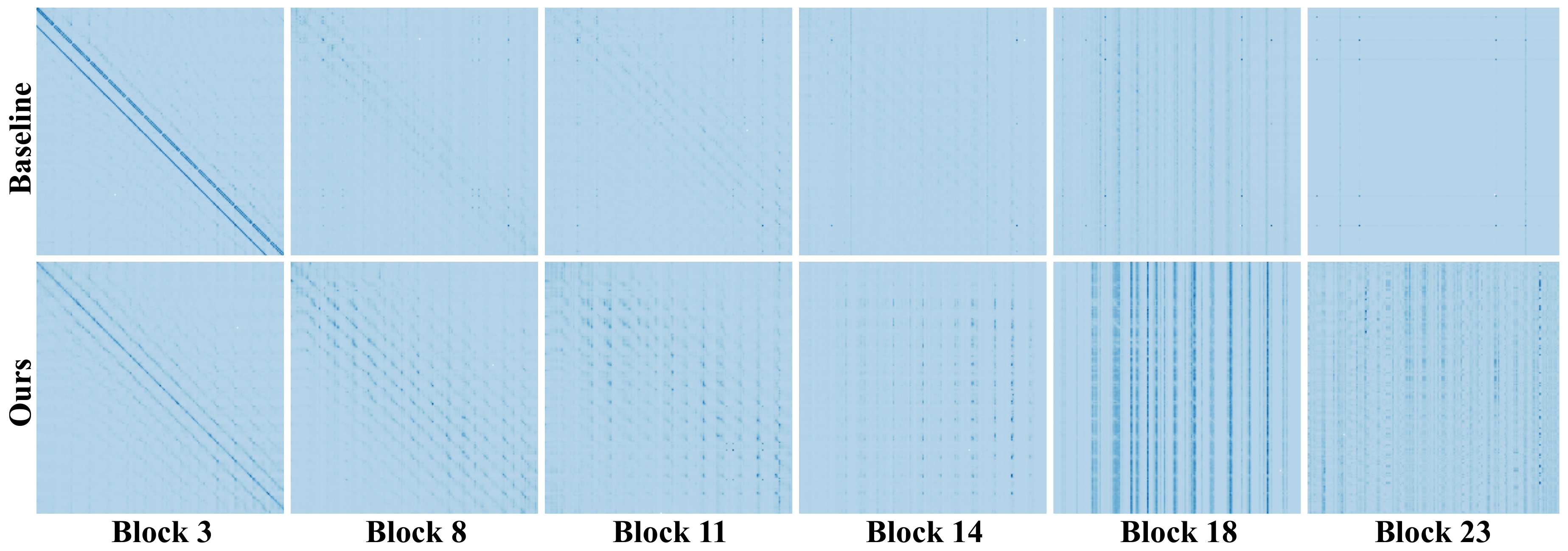}
    \vspace{-4mm}
    \caption{Comparison of the attention map (i.e., affinity matrix $\boldsymbol{A}$) between DeiT baseline ($1$st row) and our diversified variant ($2$nd row).}
    \vspace{-4mm}
    \label{fig:raw_attention}
\end{figure*}

\begin{figure}[!ht]
    \centering
    \includegraphics[width=1.0\linewidth]{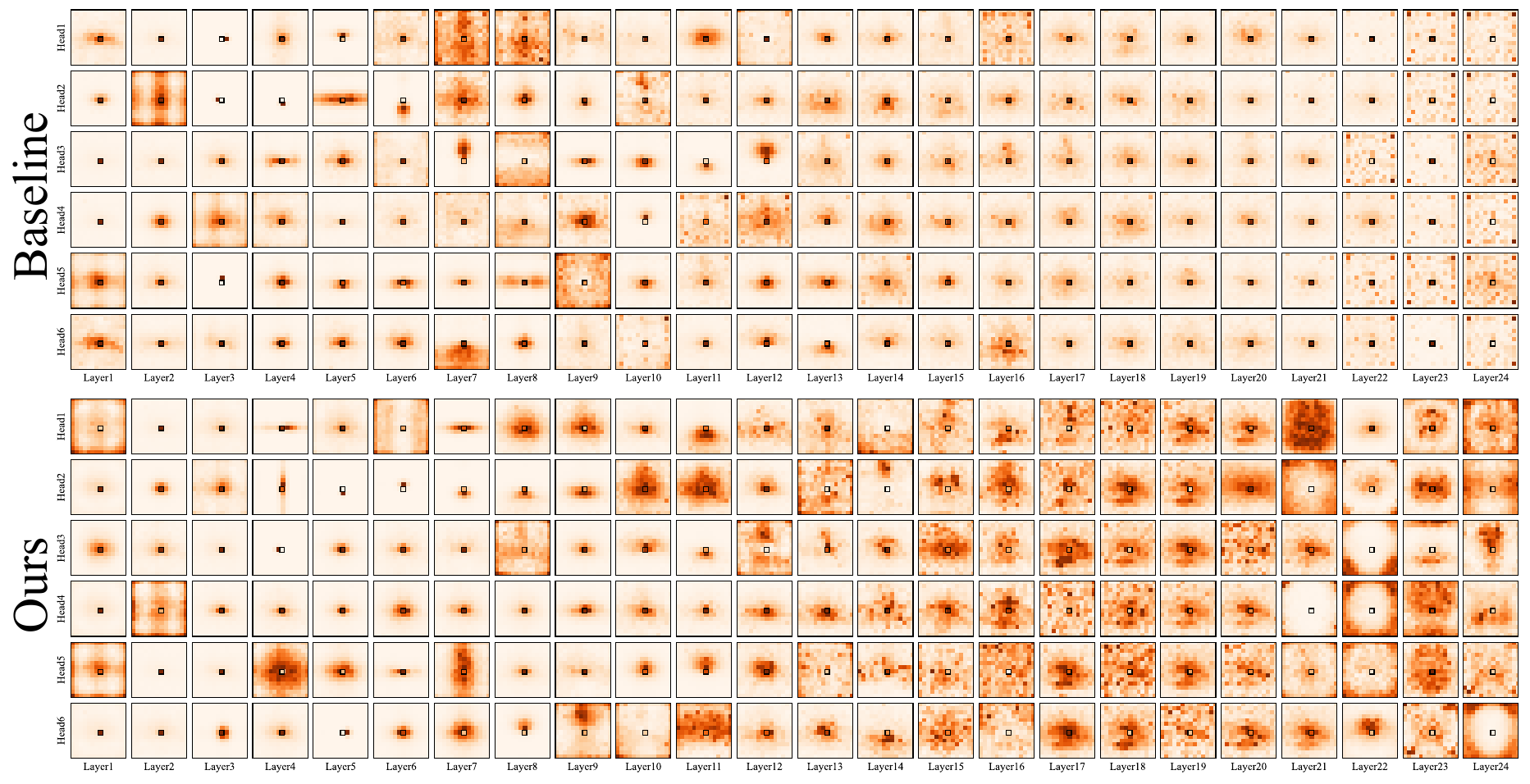}
    \vspace{-8mm}
    \caption{Attention probabilities for DeiT baseline ($1$st row) and our diversified variant ($2$nd row) with $24$ layers (columns) and $6$ heads (rows), visualized by~\cite{Cordonnier2020On}. Attention maps are averaged over $100$ test images from ImageNet. The black square is the query pixel. Zoom-in for better visibility.}
    \vspace{-4mm}
    \label{fig:prob_attention}
\end{figure}

\begin{figure}[!ht]
    \centering
    \includegraphics[width=0.93\linewidth]{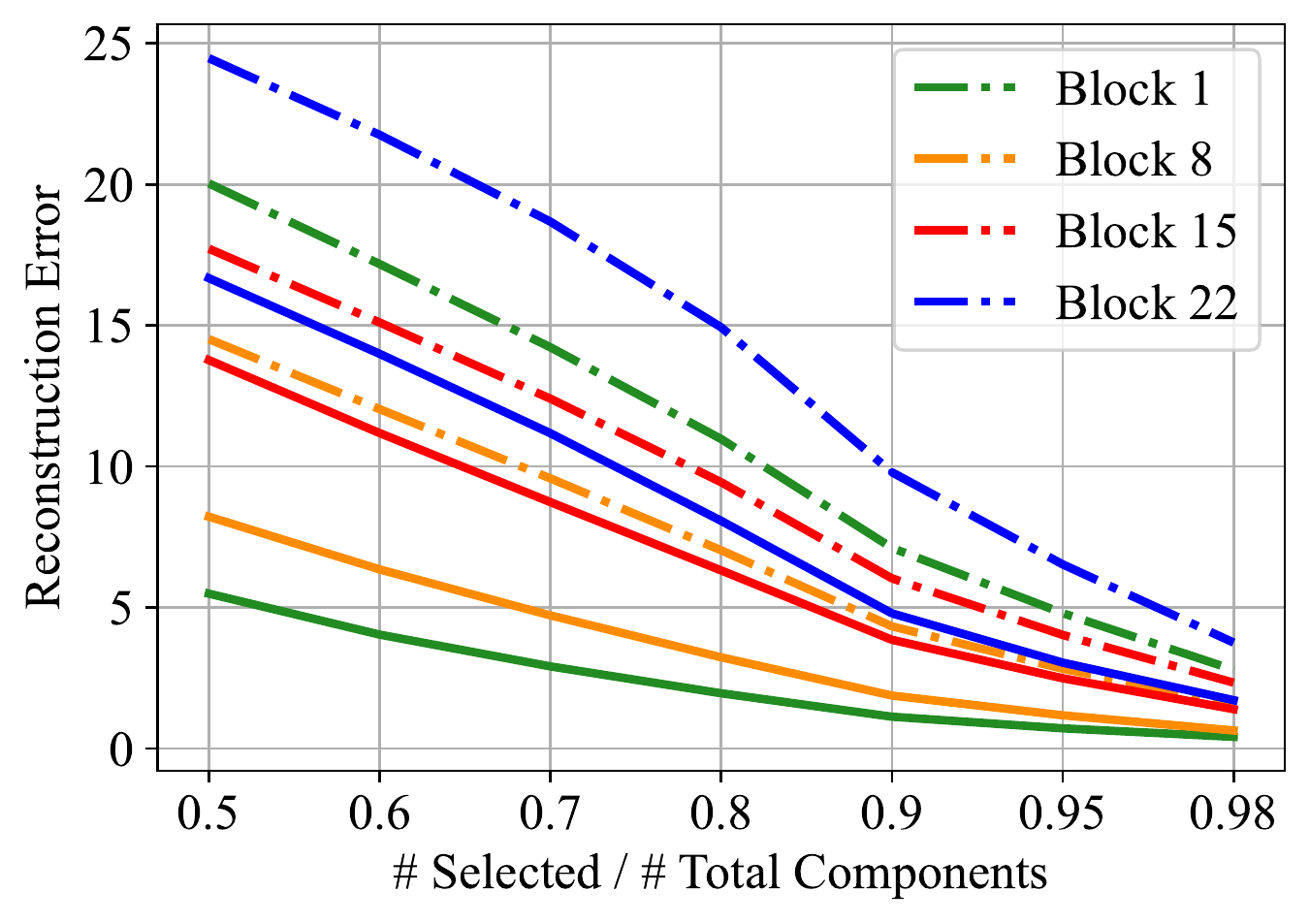}
    \vspace{-5mm}
    \caption{The weight reconstruction errors over the number of selected principle components. The dash and solid lines represent our diverse ViT and its vanilla version, respectively. The smaller error suggest a larger redundancy in the original weight space.}
    \vspace{-5mm}
    \label{fig:rank_weight}
\end{figure}

\begin{figure}[!ht]
    \centering
    \vspace{-2mm}
    \includegraphics[width=0.93\linewidth]{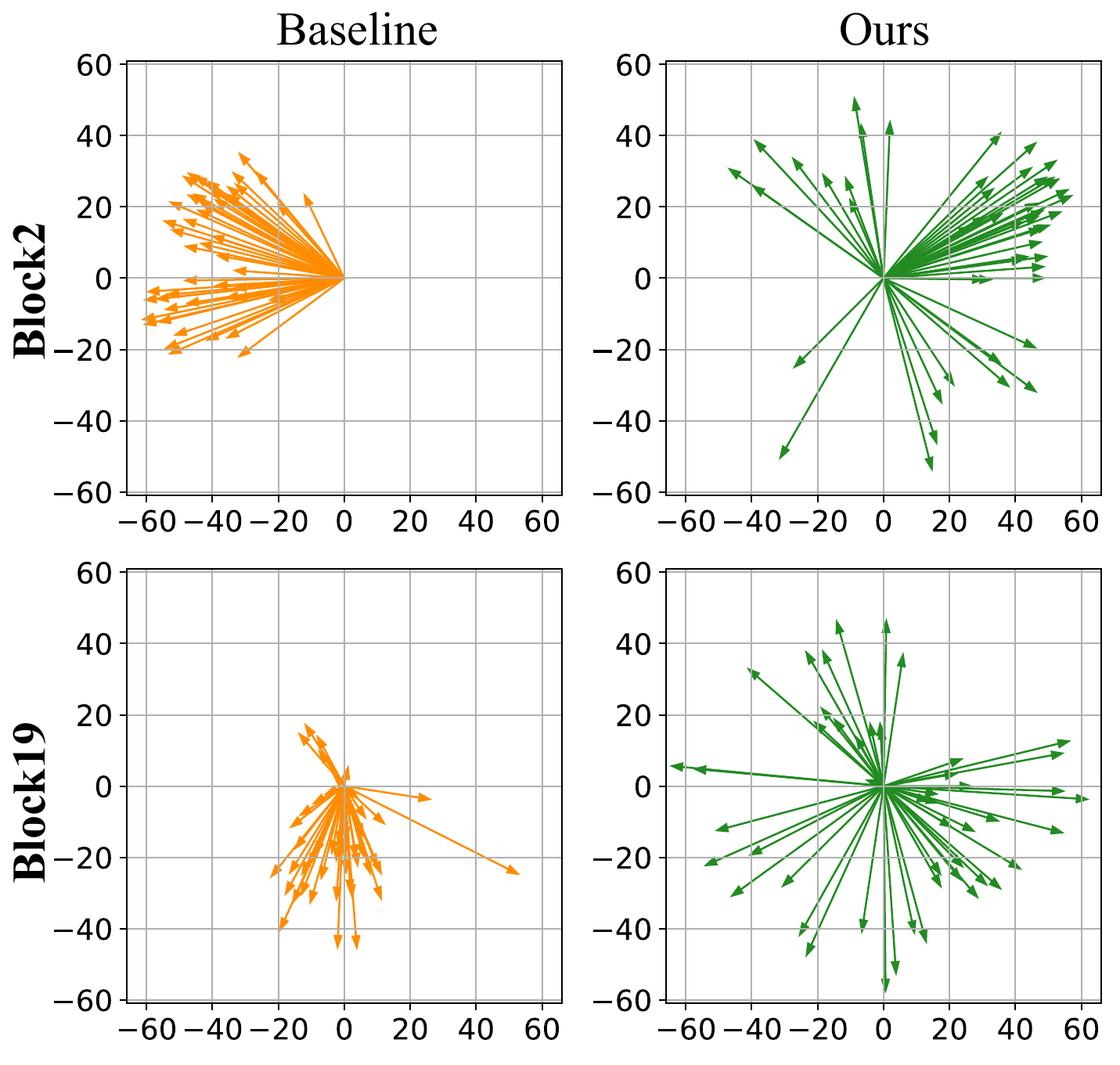}
    \vspace{-4mm}
    \caption{t-SNE of weight principle components (top-$50$) from Deit baseline ($1$st col.) and our diversified variant ($2$nd col.).}
    \vspace{-6mm}
    \label{fig:pca_weight}
\end{figure}

\paragraph{Effectively reduced redundancy.} In order to further validate the superiority of our diverse ViTs, we provide extensive qualitative and quantitative visualizations for all levels of patch embedding, attention maps, and model weights. 

\noindent \ding{182} \textit{The patch embedding level.} As presented in Figure~\ref{fig:current_vit} (\textit{right}) and Figure~\ref{fig:emb_cross}, our diverse DeiT obtains significantly reduced similarity for both within-layer and cross-layer embedding, compared to the baseline DeiT. Figure~\ref{fig:tSNE_feature} visualizes the all patch embedding of $196$ tokens for randomly selected images from ImageNet. Our methods show a 
much diverse embedding distribution and improved discriminative power. Specifically, the patch \ding{172} and \ding{173} from the left image are belong two distinct objects (i.e., human hands and dog toys) and also have totally different visual contents. However, the baseline ViT produces highly correlated patch embedding (almost fully overlapped points in Figure~\ref{fig:tSNE_feature}) and fails to distinguish them. Fortunately, these embedding redundancy and semantic ambiguity are well addressed by our diversity-aware ViT training. Similar observations can be drawn from patch pairs (\ding{174},\ding{175}), (\ding{174},\ding{176}), (\ding{177},\ding{181}), and (\ding{179},\ding{180}).

\noindent \ding{183} \textit{The attention map level.} From Figure~\ref{fig:current_vit} (right), Figure~\ref{fig:atten_cross}, and Figure~\ref{fig:atten_std}, our proposals consistently reach an enhanced diversity in terms of within-layer/cross-layer attention heads and the standard deviation within attention maps. Furthermore, we offer illustrative visualizations to show the improved representation flexibility of ViTs. Figure~\ref{fig:gradcam_attention} is generated by an advanced ViT interpretable approach~\cite{chefer2021transformer}. The heatmaps from our methods display more diverse and accurate focuses. Specifically, ours concentrate attention on more focus points that spread across the whole objects like heads, legs, and main bodies. Figure~\ref{fig:raw_attention} visualizes the attention affinity matrices $\boldsymbol{A}$. We find that the attention maps from our diverse ViT training become less uniform compared to the baseline ViT, and show stronger local patterns, which provides another possible interpretation of our beneficial diversity regularization. In the end, Figure~\ref{fig:prob_attention} dissects the fine-grained attention behaviors of certain query pixels, in which our approach shows more heterogeneous attention responses, especially in the ViT's later layers. 

\noindent \ding{184} \textit{The weight level.} As shown in Fig.~\ref{fig:current_vit} (right) and Fig.~\ref{fig:rank_weight}, given a fixed number of selected principal components, the diversified ViTs have significantly larger reconstruction errors across almost all transformer layers, implying eliminated weight redundancy and improved representative power. Consistent observations are demonstrated in Fig.~\ref{fig:pca_weight}, where our diverse ViT can widely span its weight principle components on the 2-dim space.

\begin{table}[t]
\centering
\caption{Ablation on the different combination of diversity regularizers. Experiments are conducted on DeiT-Small with ImageNet, and top-1 testing accuracy (\%) is reported.}
\label{tab:ablation_combo}
\vspace{-2mm}
\scalebox{0.7}
{\small
\begin{tabular}{c|c|cc|c|c|c}
Ablation on $\downarrow$
& \scalebox{0.75}{\rotatebox{60}{Other - Mixing Loss}}
& \scalebox{0.75}{\rotatebox{60}{Within-layer Embedding}}
& \scalebox{0.75}{\rotatebox{60}{Cross-layer Embedding}}
& \scalebox{0.75}{\rotatebox{60}{Attention Maps}}
& \scalebox{0.75}{\rotatebox{60}{Weight}}
& \rotatebox{0}{Accuracy}  \\
    
\toprule
none: DeiT-Small  & \xmark & \xmark & \xmark & \xmark & \xmark & 79.78 \\
\midrule
\multirow{5}{*}{\rotatebox{0}{\makebox{\begin{minipage}{2cm}\centering Diversity \\ Regularization \end{minipage}}}}
& \cmark & \xmarkg & \xmarkg & \xmarkg &  \xmarkg  & 79.99 \\ 
& \cmark & \cmark & \xmarkg & \xmarkg &  \xmarkg  & 80.12 \\ 
& \cmark & \xmarkg & \cmark  & \xmarkg &  \xmarkg  & 80.13 \\ 
& \cmark & \cmark & \cmark  & \xmarkg &  \xmarkg  & 80.43 \\ 
& \cmark & \cmark & \cmark  & \cmark &  \xmarkg  & 80.53 \\ \midrule
All levels diversity& \cmark & \cmark & \cmark  & \cmark &  \cmark  & \textbf{80.61} \\ 
\bottomrule
\end{tabular}}
\vspace{-5mm}
\end{table}

\subsection{Ablation Study} \label{sec:ablation}

\paragraph{Multi-level v.s. single-level diversity.} To verify the effects of diversity regularizers at different levels, we conduct an incremental evaluation on DeiT-Small with ImageNet. Achieved results are included in Table~\ref{tab:ablation_combo}. We observe that: ($i$) Diversifying patch embedding brings the most accuracy benefits ($\sim0.4\%$) and the other two levels contribute similarly ($\sim0.1\%$). It is worthy of mentioning that if adopting within-layer or cross-layer embedding diversity regularization only gains $\sim0.1\%$ accuracy while combining them leads to extra performance boosts. ($ii$) Reducing tri-level redundancy in patch embedding, attention maps, and weights establishes superior performance, which validates the effectiveness of multi-level diversity compared to single-level diversity. ($iii$) The previous useful data-level diversification, i.e., mixing loss from~\cite{gong2021vision}, can be easily plugged in our training framework and plays a complementary role in improving ViT's generalization ability. More details about the mixing loss are referred to Section~\ref{sec:more_technical}.  

\begin{table}[t]
\centering
\caption{Ablation on different categories of diversity regularization. Experiments are conducted on DeiT-Small with ImageNet, and top-1 testing accuracy (\%) is reported.}
\label{tab:ablation_single}
\vspace{-2mm}
\scalebox{0.9}
{\small
\begin{tabular}{c|l|ccc}
\toprule
\multicolumn{2}{c|}{\scalebox{1}{DeiT-Small (79.78)}} & \multicolumn{1}{c}{Weight} & \multicolumn{1}{c}{Attention} & \multicolumn{1}{c}{Embedding} \\ \midrule
\multirow{2}{*}{\rotatebox{0}{\makebox{\begin{minipage}{2cm}\centering Similarity \\ Regularization \end{minipage}}}} 
& Consine & \multirow{2}{*}{N.A.} & $79.95$ & $\textbf{80.20}$ \\ 
& Contrastive & & $69.98$ & $80.11$ \\ 
\midrule

\multirow{2}{*}{\rotatebox{0}{\makebox{\begin{minipage}{2cm}\centering Uniformity \\ Regularization \\ ~\end{minipage}}}}
& MHS & $\textbf{80.05}$ & \multirow{2}{*}{N.A.} & $80.10$ \\ 
& MGD & $79.96$ & & $79.92$ \\  \midrule   

\multirow{2}{*}{\rotatebox{0}{\makebox{\begin{minipage}{2cm}\centering Orthogonality \\ Regularization \\ ~\end{minipage}}}}
& CondO & $80.01$ & $79.90$ & $79.96$  \\ 
& SO & $79.80$ & $\textbf{80.03}$ & $80.09$ \\ 
\bottomrule
\end{tabular}}
\vspace{-5mm}
\end{table}

\vspace{-1mm}
\paragraph{How to choose different diversity regularizations.} As discussed in Section~\ref{sec:methods}, there are various regularizer options for different levels in our diverse ViTs. We implement comprehensive comparisons among these six kinds of regularizations. N.A. denotes that it is not applicable in certain levels because of unmatched design motivations. From Table~\ref{tab:ablation_single}, \underline{first}, we find that constraining layerwise attention maps in a contrastive manner might be too aggressive to hurt the performance. While patch embedding is more amenable to diversification, all examined regularizers at the embedding level boost the ViT's performance. \underline{Secondly}, compared to strict orthogonality regularization, the general hypersphere uniformity is more preferred by the weight level of ViTs. \underline{Lastly}, at the attention level, SO outperforms all other choices. A possible reason is that orthogonal attention maps not only enjoy an unleashed representation flexibility but also stabilize and smoothen ViT training due to reduced Lipschitz constant, as suggested by~\cite{zhang-etal-2021-orthogonality}.

\vspace{-1mm}
\paragraph{Training time analyses.} Since the computation of our regularizers is quite cheap (e.g., $2$ steps power iteration), the extra time overhead is moderate, as indicated below of per epoch training time on eight Quadro RTX 6000 GPUs.

\begin{table}[!ht]
\centering
\vspace{-3mm} \label{tab:time}
\resizebox{\linewidth}{!}{
\begin{tabular}{@{}l|ccccc}
\toprule
Settings & Baseline & + Embedding Reg. & + Attention Reg. & + Weight Reg. & + All Reg. \\ \midrule
Training Time (s) & 582 & 590 & 595 & 677 & 695    \\
\bottomrule
\end{tabular}}
\vspace{-3mm}
\end{table}%

\vspace{-1mm}
\section{Conclusion and Broader Impact} 
In this paper, we, for the first time, systematically reveal the broad existence of redundancy at all token embedding, attention map, and weight levels in vision transformers, which limits the ViT's expressiveness and flexibility. We address this issue following the principle of diversity during ViT training. Comprehensive experiments on ImageNet across diverse ViT backbones demonstrate that equipped diversity regularizers effectively eliminate the redundant representation and lead to superior generalization.

For the limitation of this work, we only focus on vision transformers rather than general transformers in natural language processing, which we leave for future works. Meanwhile, although our paper is scientific in nature, it might amplify the existing societal risk of applying ViTs since we have no control of anyone who can get access to our improved training algorithms. A potential solution is to issue licenses and limit the abuse.

\clearpage

{\small
\bibliographystyle{ieee_fullname}
\bibliography{DV}
}

\clearpage

\appendix

\renewcommand{\thepage}{A\arabic{page}}  
\renewcommand{\thesection}{A\arabic{section}}   
\renewcommand{\thetable}{A\arabic{table}}   
\renewcommand{\thefigure}{A\arabic{figure}}

\section{More Technical Details} \label{sec:more_technical}

\paragraph{Token-wise mixing loss.} As an effective data-level diversification from~\cite{gong2021vision}, it mixes the input patches from two different images and leverages an additional shared classifier to output patch embedding for the classification of each patch. It can be described as follows:
\begin{equation}
    \mathcal{R}_{\mathrm{mixing}}(\mathbf{x}) = \frac{1}{n} \sum_{i=1}^n \mathcal{L}_{\mathrm{XE}}(g(\boldsymbol{e}_i^{\mathrm{L}}),y_i),
\end{equation}
where $\boldsymbol{e}_i^{\mathrm{L}}$ is the patch representation in the last layer, $g$ represents the additional shared linear classifier, $y_i$ denotes the label of the corresponding patch, and $\mathcal{L}_{\mathrm{XE}}$ stands for the cross entropy loss. 

\section{More Implementation Details} \label{sec:more_implementation}

\paragraph{Hyperparameters of our diversity regularizers.} Table~\ref{tab:hyperparamters} summarizes our adopted hyperparameters during diversity-aware ViT training.

\begin{table}[!ht]
\centering
\caption{Detailed hyperparamters of our diversity regularization.}
\label{tab:hyperparamters}
\vspace{-2mm}
\scalebox{0.75}
{\small
\begin{tabular}{l|ccccc}
\toprule
 \multirow{2}[2]{*}{Settings} & \multirow{2}[2]{*}{Mixing loss} & \multirow{2}[2]{*}{Weight} & \multirow{2}[2]{*}{Attention} & \multicolumn{2}{c}{Embedding} \\ 
  & & & & Within-layer & Cross-layer \\\midrule
ViT-Small & $1$ & $5\times 10^{-4}$ & $1\times10^{-4}$ & $0.5$ & $0.5$ \\  
ViT-Base & $1$ & $5\times 10^{-5}$ & $1\times10^{-5}$ & $0.5$ & $0.5$ \\  
DeiT-Small & $1$ & $5\times 10^{-4}$ & $1\times10^{-4}$ & $0.5$ & $0.5$ \\ 
DeiT-Small24 & $1$ & $5\times 10^{-4}$ & $1\times10^{-4}$ & $0.5$ & $0.5$ \\ 
DeiT-Base & $1$ & $1\times 10^{-6}$ & $5\times 10^{-6}$ & $0.5$ & $0.5$ \\ 
Swin-Small & $1\times 10^{-3}$ & $1\times 10^{-6}$ & $1\times 10^{-3}$ & $0.9$ & -   \\ 
Swin-Base &  $1$ & $1\times 10^{-6}$ & $1\times 10^{-3}$ & $0.5$ & -\\ 
\bottomrule
\end{tabular}}
\vspace{-4mm}
\end{table}

\section{More Experiment Results} \label{sec:more_results}

\begin{figure}[!ht]
    \centering
    \includegraphics[width=1.0\linewidth]{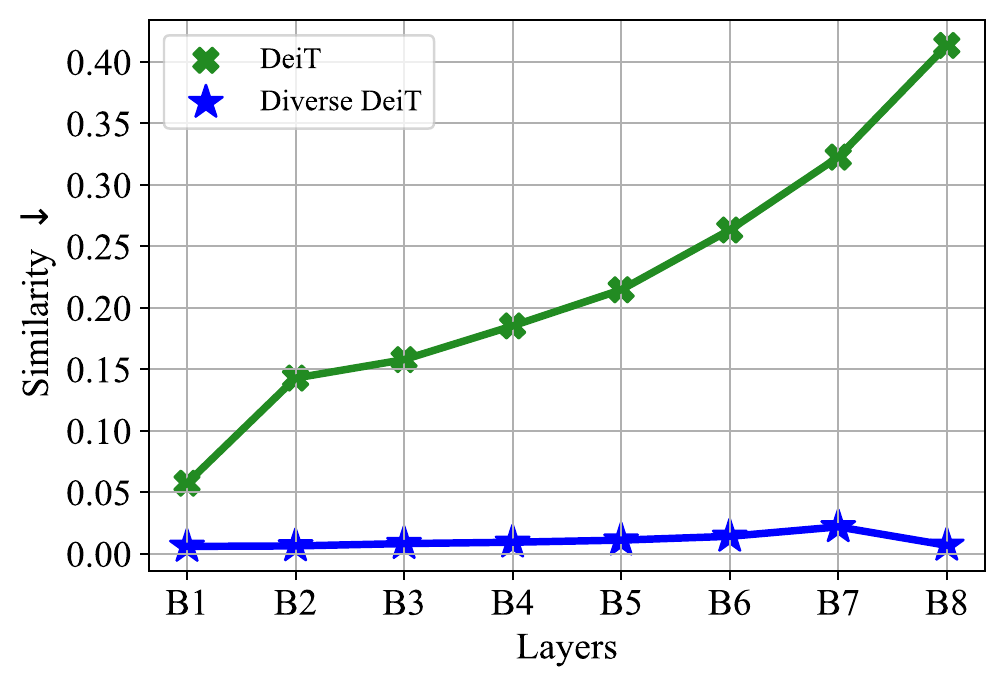}
    \vspace{-8mm}
    \caption{The cross-layer \textbf{patch embedding similarity} of DeiT-Small and its diversified version on ImageNet. We calcluate the \textit{cosine} similarity between embedding from each layer and the final layer. The smaller number indicates better diversity.}
    \vspace{-2mm}
    \label{fig:emb_cross}
\end{figure}

\paragraph{Cross-layer diversity on patch embedding.} Figure~\ref{fig:emb_cross} shows that our methods substantially shrink the similarity of cross-layer patch embedding. 

\begin{figure}[!ht]
    \centering
    \includegraphics[width=1.0\linewidth]{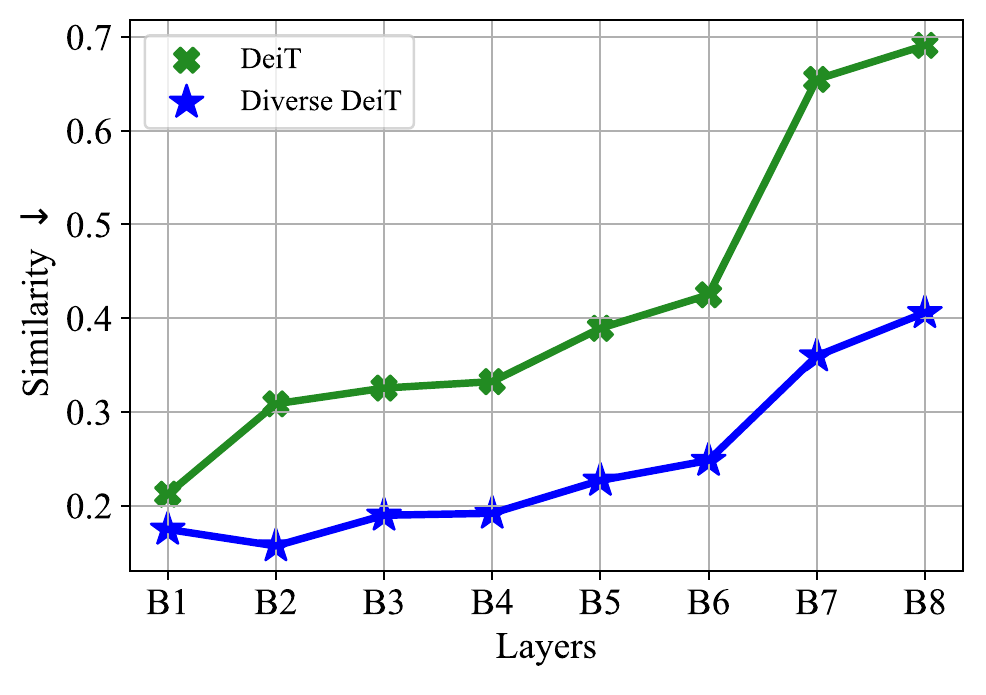}
    \vspace{-8mm}
    \caption{The cross-layer \textbf{attention maps similarity} of DeiT-Small and its diversified version on ImageNet. We calcluate the \textit{cosine} similarity between embedding from each layer and the final layer. The smaller number indicates better diversity.}
    \vspace{-2mm}
    \label{fig:atten_cross}
\end{figure}

\paragraph{Cross-layer diversity on attention.} As shown in Figure~\ref{fig:atten_cross}, our diverse ViT obtains a consistently lower correlation of cross-layer attention maps. 

\begin{figure}[!ht]
    \centering
    \includegraphics[width=1.0\linewidth]{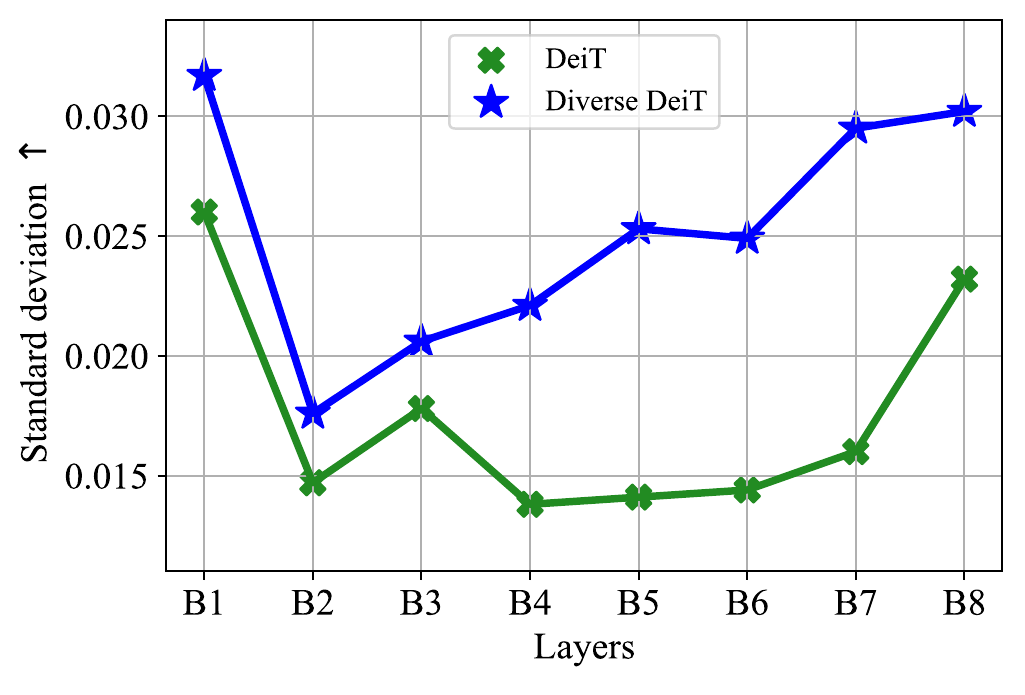}
    \vspace{-8mm}
    \caption{The averaged standard deviation within attention maps of DeiT-Small and its diversified version on ImageNet. The larger number indicates better diversity.}
    \vspace{-2mm}
    \label{fig:atten_std}
\end{figure}

\paragraph{Standard deviations within attention maps.} From Figure~\ref{fig:atten_std}, we observe that the averaged standard deviations within attention maps are amplified by our approaches, suggesting an enhanced diversity. Note that we do not explicitly regularize the standard deviations of attention. 

\end{document}